\pdfoutput=1
\documentclass[11pt]{article}

\usepackage[preprint]{coling}

\usepackage{times}
\usepackage{latexsym}

\usepackage[T1]{fontenc}

\usepackage[utf8]{inputenc}

\usepackage{microtype}

\usepackage{inconsolata}

\usepackage{graphicx}

\usepackage{hyperref}
\usepackage{url}
\usepackage{multicol}
\usepackage{multirow}
\usepackage{booktabs}
\usepackage{array}
\usepackage{amsmath}
\usepackage{subcaption}
\usepackage{float}
\usepackage{enumitem} 
\usepackage{pifont}
\usepackage{twemojis}
\usepackage{txfonts}
\usepackage{amssymb}
\usepackage{subcaption}
\usepackage{algorithm}
\usepackage{algpseudocode}
\usepackage[english]{babel}

\definecolor{Increase}{HTML}{D0F4DE}
\definecolor{Decrease}{HTML}{FEC8C3}

\title{\textbf{MEOW}: \textbf{MEMO}ry Supervised LLM Unlearning Via Inverted Facts}

\author{
 \textbf{Tianle Gu\textsuperscript{\twemoji{1f4dd},\twemoji{1f431}}\thanks{{}{} Work done during internship at Shanghai Artificial Intelligence Laboratory.}\thanks{{}{} Equal Contribution}},
 \textbf{Kexin Huang\textsuperscript{\twemoji{1f408}}$\footnotemark[2]$},
 \textbf{Ruilin Luo\textsuperscript{\twemoji{1f4dd}}},
 \textbf{Yuanqi Yao\textsuperscript{\twemoji{1f431}}},
\\
 \textbf{Yujiu Yang\textsuperscript{\twemoji{1f4dd}}\thanks{{}{} Corresponding Authors}},
 \textbf{Yan Teng\textsuperscript{\twemoji{1f431}}$\footnotemark[3]$},
 \textbf{Yingchun Wang\textsuperscript{\twemoji{1f431}}}
\\
 \textsuperscript{\twemoji{1f4dd}}Tsinghua Shenzhen International Graduate School, Tsinghua University, \\
 \textsuperscript{\twemoji{1f431}}Shanghai Artificial Intelligence Laboratory,
 \textsuperscript{\twemoji{1f408}}Fudan University
\\
}

\begin{document}
\maketitle
\begin{abstract}
Large Language Models~(LLMs) can memorize sensitive information, raising concerns about potential misuse.
LLM Unlearning, a post-hoc method to remove this information from trained LLMs, offers a promising solution to mitigating these risks.
However, previous practices face three key challenges: \textit{1. Utility}: successful unlearning often causes catastrophic collapse on unrelated tasks.
\textit{2. Efficiency:} many methods either involve adding similarly sized models, which slows down unlearning or inference, or require retain data that are difficult to obtain.
\textit{3. Robustness}: even effective methods may still leak data via extraction techniques.
To address these challenges, we propose \textsc{MEOW}, a simple yet effective gradient descent-based unlearning method.
Specifically, we use an offline LLM to generate a set of inverted facts. Then, we design a new metric, \textsc{MEMO}, to quantify memorization in LLMs. Finally, based on the signals provided by MEMO, we select the most appropriate set of inverted facts and finetune the model based on them.
We evaluate \textsc{MEOW} on the commonly used unlearn benchmark, ToFU, with Llama2-7B-Chat and Phi-1.5B, and test it on both NLU and NLG tasks. 
Results demonstrate significant improvement of \textsc{MEOW} in forget quality without substantial loss in model utility. 
Meanwhile, \textsc{MEOW} does not exhibit significant drop in NLU or NLG capabilities, and there is even a slight increase in NLU performance. \footnote[1]{{} {}Codes and data are available at \href{https://github.com/Carol-gutianle/MEOW.git}{Github}.}
\end{abstract}

\section{Introduction}
\begin{figure*}[!t]
    \centering
    \includegraphics[width=1\linewidth]{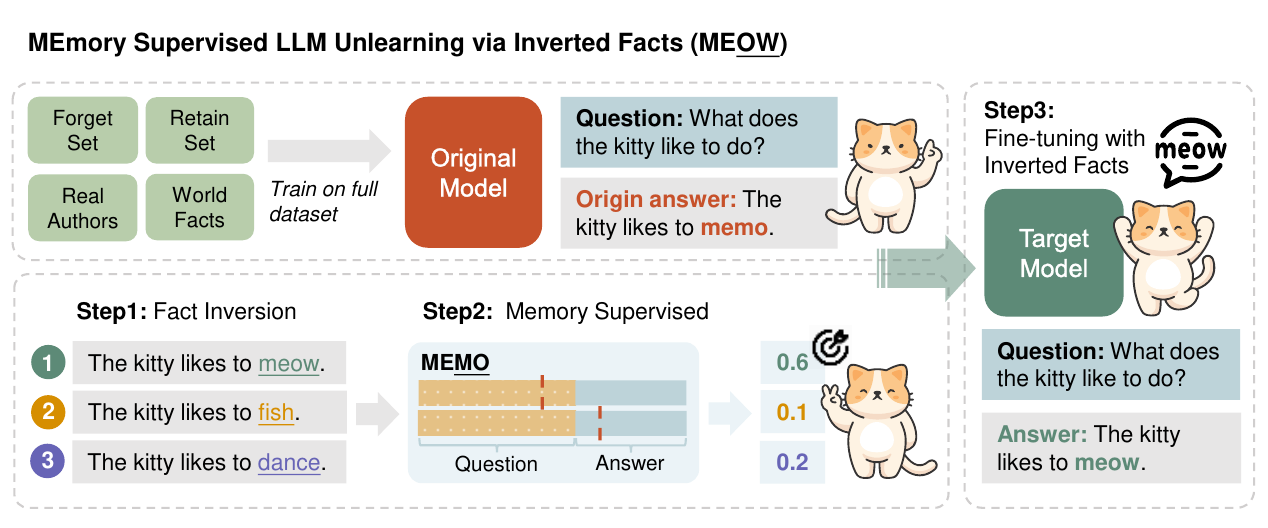}
    \caption{Overview of MEOW.}
    \label{fig:overflow}
\end{figure*}

Recent research ~\cite{Hartmann2023SoKMI,tirumala2022memorizationoverfittinganalyzingtraining} highlights that LLMs have the potential to memorize training data, which can be exposed through red teaming attacks~\citep{nasr2023scalableextractiontrainingdata} like Membership Inference Attack~(MIA)~\citep{shokri2017membership,shi2024detecting} and Prompt Injection~\citep{Khomsky2024PromptIA}.
Such vulnerabilities raise concerns about privacy leakage and copyright violations. For instance, in medical LLMs, malicious users could extract training data to guess whether a patient has a specified disease.
Meanwhile, unintended data leakage, without the awareness or consent of data owners, may result in violations of related laws, such as the General Data Protection Regulation~\citep{gdpr} in the European Union.

So, how to protect sensitive information from potential leakage?
Data pre-processing~\citep{10.1145/1179601.1179608,dernoncourt2016deidentificationpatientnotesrecurrent,lison-etal-2021-anonymisation,kandpal2022deduplicatingtrainingdatamitigates,ghosh2024aegisonlineadaptiveai} and Differential Privacy~(DP)~\citep{dwork2006calibrating,dwork2008differential,Abadi_2016,anil2021largescaledifferentiallyprivatebert,li2022largelanguagemodelsstrong,yu2022differentiallyprivatefinetuninglanguage} are widely studied and established to prevent data leakage.
Data pre-processing involves data audit and removing all sensitive information from training data, while DP adds random noise to data, making sensitive and normal information indistinguishable. 
However, data pre-processing requires numerous annotations, and both approaches necessitate retraining the model – an impractical solution for LLMs.

\label{sec:intro}
Therefore, applied in a post-processing manner, LLM unlearning offers a promising solution.
Based on the access of the model, previous research can be divided into three schools of thought:
\ding{182}~\textbf{Black Box Setting~(BBS)}, where model weights are totally inaccessible. Approaches under this setting are often inference-based, such as In-Context-Learning~(ICL;~\citet{pawelczyk2024incontextunlearninglanguagemodels}).
\ding{183}~\textbf{Grey Box Setting~(GBS)}, where partial access to the model is available, such as logits or embedding space. Approaches under this setting are always input-~\citep{liu2024largelanguagemodelunlearning} or output-based~\citep{huang2024offsetunlearninglargelanguage,ji2024reversingforgetretainobjectivesefficient}.
\ding{184}~\textbf{White Box Setting~(WBS)}, where the full model weights are accessible. Under this setting, approaches are typically based on fine-tuning~(e.g., Gradient Ascent~\citep{yao2024largelanguagemodelunlearning} and its variants), preference optimization~\citep{rafailov2024directpreferenceoptimizationlanguage,zhang2024negativepreferenceoptimizationcatastrophic}, knowledge distillation~\citep{wang2024rkldreversekldivergencebasedknowledge}, and model editing~\citep{wang2024largescaleknowledgewashing}.

\noindent Although previous practices have facilitated effective unlearning to some extent, it remains essential to critically reassess them from three perspectives.
First, \textit{do these approaches successfully maintain model utility?} WBS approaches often encounter catastrophic forgetting of content that does not require unlearning. This issue is particularly pronounced in Gradient Ascent~(GA)-based methods, where unbounded loss divergence exists as a significant issue~\citep{zhang2024negativepreferenceoptimizationcatastrophic}.
Second, \textit{the efficiency of these methods counts}. Generally, efficiency is evaluated from two aspects: At the model level, methods such as preference optimization, knowledge distillation~(KL)-based, and logits ensemble-based approaches often require a proxy model of equal or smaller size for assistance, which slows down training or inference.
At the data level, some methods depend on access to the retain dataset to maintain the model's utility. However, obtaining the entire retain dataset is nearly impossible; otherwise, it would be feasible to simply retrain a model from scratch.
Finally, \textit{can the unlearned model be re-exploited to recover the forgotten data, i.e., does the method possess robustness?} Such issues often arise with the BBS and GBS methods. If the origin model is a white-box model, attackers can still reproduce the forgotten data if they obtain the complete weights.

To tackle these challenges, we propose an easy yet effective approach, MEOW, simultaneously considering utility, efficiency, and robustness.
Under WBS, MEOW is a gradient descent-based method that avoids loss divergence and eliminates the need for auxiliary models or retain datasets. 
It modifies the model's weights to unlearn target data, after which the modified model can be safely open-sourced while preventing attackers from extracting the removed information, ensuring the robustness of unlearning.
Fig.~\ref{fig:overflow} illustrates our workflow.
In detail, we argue that accurately quantifying the memorization of sensitive information is the first step toward effective unlearning, in LLMs stems from this memorization. To address this, we introduce a novel metric, MEMO, to measure the memorization of individual/group sequences in LLMs. 
Next, we generate a set of alternative answers based on undesired responses from the forgetting dataset. Guided by MEMO's memorization signals, we select the largest/smallest $k$ memorized answers as labels to form a perturbation dataset. Finally, we fine-tune the origin model on this dataset. Extensive experiments, on the unlearning, NLG, and NLU benchmarks, demonstrate the superior performance over existing methods of MEOW.

We summarize our contributions as follows:

\noindent$\bullet$ We propose MEMO, a novel metric for quantifying memorization in LLMs, offering superior effectiveness, efficiency, and compatibility with MEOW compared to traditional methods.

\noindent$\bullet$ Our simple yet effective method, MEOW, shows a significant improvement in forget quality without causing a substantial decline in model utility. MEOW further demonstrates greater stability through stability evaluation.

\noindent$\bullet$ Extensive experiments on NLU and NLG datasets show that MEOW preserves models' original capabilities, with NLU performance even improving on some datasets after unlearning.
\section{Settings, Goals, and Evaluation}
\subsection{Settings}
Suppose we have a dataset $D=(x,y)$ and an untrained LLM $M_\text{u}$. After training $M_\text{u}$ on $D$, we obtain a trained LLM, $M_\text{o}$, which serves as the original model for the unlearning task.
Meanwhile, we divide the dataset into $D_\text{f}=(x^f, y^f)$ and $D_\text{r}=(x^r,y^r)$, representing the dataset to forget and the dataset to retain.
We train $M_\text{u}$ on $D_\text{r}$ to obtain the retain model $M_\text{r}$ as the ground truth for unlearning tasks.
Furthermore, we introduce an additional dataset $D_\text{g}=(x^g,y^g)$ to evaluate the general capabilities of the model after unlearning, such as its NLU and NLG abilities. 

\subsection{Goals}
After unlearning, the origin model $M_\text{o}$ is transformed into the target model $M_\text{t}$. We categorize the unlearning goals into hard unlearning and soft unlearning, based on the format of responses $\Tilde{y}$ that $M_\text{t}$ generates to prompts in $D_\text{f}$. 
Hard unlearning refers to responses where the target model $M_\text{t}$ either avoids answering, providing blank or template answers like ``I don't know'', or generates completely nonsensical responses. 
Soft unlearning, however, involves providing incorrect but understandable answers.
For general-purpose LLMs, hard unlearning would greatly harm the user experience. Therefore, soft unlearning is more suitable for ideal LLM unlearning, which is the goal of our paper.
We discuss the potential limitations in Sec.~\ref{sec:limitation}.

\subsection{Evaluation}
Nearly all the LLM unlearning algorithms are trying to address the problem of balancing model utility and forget quality, i.e., how to effectively unlearn without causing catastrophic collapse on $D_\text{r}$ and $D_\text{g}$. 
Therefore, this paper utilizes 2 metrics: 
\ding{182}~\textbf{Model utility}: typically measures the performance of $M_\text{t}$ on $D_\text{r}$ and $D_\text{g}$.
\ding{183}~\textbf{Forget quality}: can be assessed in two ways, measuring the difference between $M_\text{t}$ and $M_\text{o}$ on $D_\text{f}$, or the similarity between $M_\text{t}$ and $M_\text{r}$.
For the former way, both hard and soft unlearning can achieve high forget quality. However, for the latter,
hard unlearning typically fails to maintain high forget quality due to its negative impact on model utility.
Therefore, we believe the latter one is more rigorous and aligns better with real-world scenarios, and use it for the measurement of forget quality.

\section{Methodology}
\subsection{Quantifying memorization in LLMs}
\label{subsec:memo}
\begin{figure}[tbp]
    \centering
    \includegraphics[width=1\linewidth]{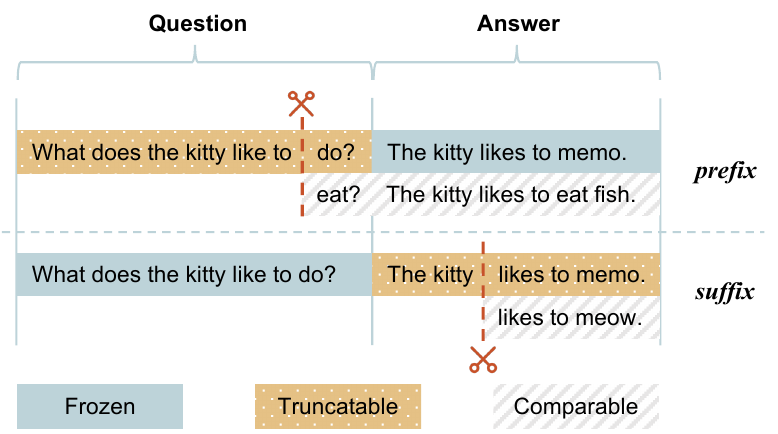}
    \caption{MEMO with \textit{prefix} or \textit{suffix} mode. }
    \label{fig:modes}
\end{figure}

\paragraph{MEMO}
Given a question $x = \{x_i \mid 0 \leq i < |x|\}$ and an answer $y = \{y_i \mid 0 \leq i < |y|\}$, we segment $x$ and $y$ according to different modes, as shown in Fig.~\ref{fig:modes}. 
Specifically, in the \textit{prefix} mode, we truncate $x$ to form prompt $T_{\text{p}} = x_{0}^{e}$, where $e$ represents the truncation endpoint.
In the \textit{suffix} mode, we truncate $y$ to form $T_{\text{p}} = x + y_{0}^{e}$. 
And the remaining part of the sequence is the ground truth $T_{\text{gt}}$ to be compared, defined as:
\begin{equation}
T_{\text{gt}} = 
\begin{cases}
x_{e+1}^{|x|} + y, & \text{if \textit{prefix} mode}, \\
x + y_{e+1}^{|y|}, & \text{if \textit{suffix} mode}. \nonumber
\end{cases}
\label{eq:t_gt}
\end{equation}
Then, we feed $T_{\text{p}}$ into the model $M$, obtaining the output $T_\text{r}$. We compare $T_{\text{r}}$ with $T_{\text{gt}}$ using Rouge, as specified in Eq.~\ref{eq:memo}:
\begin{equation}
\text{MEMO}(x, y) = \frac{\sum_{i=1}^{N} \text{Rouge-N}(T_{\text{r}}, T_{\text{gt}})}{S},
\label{eq:memo}
\end{equation}
where Rouge-N refers to the Rouge~\citep{lin-2004-rouge}, and $S$ denotes the total number of sliding windows.
Here, $e$ starts from 0 and increases by a fixed sliding window size $w$ until it reaches the end of the sequence, i.e., $e \leq |EOS|$.
The pseudocode for MEMO is provided in the App.~\ref{app:pseudo_memo}.

\paragraph{MEMO Strength}
For any dataset, we measure the memorization of a model for a certain prompt-response pair $(x, y)$ by calculating $\text{MEMO}(x, y)$ and obtain the average value, denoted as $\mu$.
\begin{align}
    \mu(D,M) &= \frac{\sum_{i=1}^{N} \mathrm{MEMO} \left( x_i, y_i \right) }{S} \nonumber
\end{align}

\paragraph{MEMO Consistency}
We introduce $c_v(D,M)$ to represent the variance of memorization in $M$ for a given sample set $D$, i.e., the consistency of memorization across different samples.

\begin{align}
    \sigma(D,M) &= \sqrt{\frac{\sum_{i=1}^{N} \left( \mathrm{MEMO}(x_i, y_i) - \mu(D,M) \right) ^ 2}{S}} \nonumber \\ 
    c_v(D,M) &= \frac{\sigma(D,M)}{\mu(D,M)} \label{eq:cv} \nonumber
\end{align}

\subsection{LLM Unlearning Via Inverted Facts}
\paragraph{Conceptual Motivation}
In our method, we build on the Information Overloaded Theory~\citep{10.5555/1213384}, which suggests that excessive information can impair normal understanding and decision-making. 
Applied to LLMs, we interpret direct exposure to specific sensitive information as a ``strong belief'' in a particular fact.
However, when presented with more similar but different or even contradictory facts, the model becomes hesitant and tends to discard the original belief.

\paragraph{Fact Inversion}
For the forgetting dataset $D_\text{f}$ and the facts that need to be forgotten, we use an offline LLM~\citep{achiam2023gpt} to generate inverted facts. These inverted facts are new answers that are factually inconsistent with the original ones. For instance, in Fig.~\ref{fig:overflow}, for the fact ``The kitty likes to memo,'' we generate three reversed facts: ``The kitty likes to meow'', ``The kitty likes to fish'', and ``The kitty likes to dance''.
We provide the prompt used for fact inversion in App.~\ref{app:prompt_used_for_fact_inversion}.

\paragraph{Memory Supervised}
\label{sec:mem_enhance}
For the generated inverted facts, we use MEMO to calculate the memorization of each fact. Then, we select the top or bottom $k$ facts with the highest or lowest memorization to form a new fact set. 
Given our primary focus on the memorization of answers, we adopt the \textit{Suffix} mode.
Additionally, for hyperparameters $w$, and $N$, which control the length of the sliding window and the choice of $\mathrm{Rouge\text{-}N}$, we use window size $w=5$ and $\mathrm{Rouge\text{-}1}$ in our experiments.

\paragraph{Fine-tuning with Inverted Facts}
Finally, we fine-tune the model using the selected inverted facts and train it with the next-token prediction task. We employ cross-entropy loss~(CE) that constrains the similarity between estimated and ground-truth tokens, which can be presented as
\begin{equation}
    L = CE(\Tilde{y}, \hat{y}), \nonumber
\end{equation}
where $\Tilde{y}$ is the predicted token, and $\hat{y}$ is the ground-truth token.
\section{Experiments}
\subsection{Baselines}
The unlearning method under the WBS can be considered as fine-tuning the original model with an unlearning objective function, which is a specific combination of the loss on the forget data and the loss on the retain data, as shown in Eq.~\ref{eq:loss}~\citep{liu2024rethinkingmachineunlearninglarge}.
The forget losses include:~\ding{182} GA~\citep{yao2024largelanguagemodelunlearning}: performs gradient ascent on forget data.
~\ding{183} DPO~\citep{rafailov2024directpreferenceoptimizationlanguage}: direct preference optimization, encouraging the model to give responses like ``I don't know''.
~\ding{184} NPO~\citep{zhang2024negativepreferenceoptimizationcatastrophic}: negative preference optimization, a variant of DPO where only the correct answer is used as a negative label. 
The retain losses include: 
~\ding{182} GD~\citep{maini2024tofutaskfictitiousunlearning,jia2024soulunlockingpowersecondorder}: subtracts the loss on forget data from the loss on retain data. ~\ding{183} KL~\citep{wang2024rkldreversekldivergencebasedknowledge,maini2024tofutaskfictitiousunlearning}: calculates the KL-divergence on retain data before and after unlearning to ensure that the model retains its original performance on retain data. We term each baseline by combining the specific forget loss and retain loss, e.g., GA+KL indicates the use of GA as the forget loss and KL as the retain loss.

\begin{align}
    \mathcal{L}_\text{f} &= \mathbb{E}_{(x, y) \in D_\text{f}} \left[\ell\left(y \mid x; \boldsymbol{\theta}\right)\right] \nonumber \\
    \mathcal{L}_\text{r} &= \mathbb{E}_{(x, y) \in D_\text{r}} \left[\ell(y \mid x; \boldsymbol{\theta})\right] \nonumber \\
    \mathcal{L} &= - {\mathcal{L}_\text{f} + \lambda \mathcal{L}_\text{r}}
    \label{eq:loss}
\end{align}
Here, $\lambda$ controls the retain strength, and $\ell(y \mid x; \boldsymbol{\theta})$ denotes the prediction loss of using $\boldsymbol{\theta}$ when given the input $x$ with respect to the response $y$.

\subsection{Experiments on Unlearning Dataset}
\begin{table*}[htbp]
    \centering
    \scalebox{0.85}{
    \begin{tabular}{ccccccccccccc}
        \toprule
        \multirow{3}{*}{\textbf{Method}} & \multicolumn{4}{c}{ToFU-1\%} & \multicolumn{4}{c}{ToFU-5\%} &\multicolumn{4}{c}{ToFU-10\%} \\  
        \cmidrule(lr){2-5} \cmidrule(lr){6-9} \cmidrule(lr){10-13}
        & \multicolumn{2}{c}{Llama 2} & \multicolumn{2}{c}{Phi-1.5} &\multicolumn{2}{c}{Llama 2} & \multicolumn{2}{c}{Phi-1.5} &\multicolumn{2}{c}{Llama 2} & \multicolumn{2}{c}{Phi-1.5}\\ 
        \cmidrule(lr){2-5} \cmidrule(lr){6-9} \cmidrule(lr){10-13}
        & M.U.  & F.Q.  & M.U.  & F.Q.  & M.U.  & F.Q.        & M.U.  & F.Q.  & M.U.  & F.Q.  & M.U.  & F.Q. \\  \midrule
        Origin Model & 0.62 & 0.00 & 0.52 & 0.00 & 0.62 & 0.00 & 0.52 & 0.00 & 0.62 & 0.00 & 0.52 & 0.00 \\
        Retain Model & 0.62 & 1.00  & 0.52 & 1.00 & 0.62 & 1.00 & 0.52 & 1.00 & 0.62 & 1.00 & 0.52 & 1.00 \\ \midrule
        
        GA & 0.52 & 0.40 & \underline{0.51} & 0.00 & 0.37 & 0.05 & 0.07 & \underline{0.14} & 0.00 & 0.00 & 0.21 & 0.00 \\
        GD & 0.53 & 0.27 & \underline{0.51} & 0.00 & 0.33 & 0.11 & \underline{0.41} & 0.00 & 0.17 & 0.00 & 0.31 & 0.03 \\
        GA+KL & 0.53 & 0.40 & 0.50 & 0.00 & 0.35 & 0.14 & 0.28 & 0.09 & 0.05 & 0.00 & 0.28 & 0.41 \\\midrule
        DPO & \textbf{0.58} & 0.27 & \textbf{0.52} & 0.00 & 0.02 & 0.00 & 0.39 &0.00 & 0.00 & 0.00 & 0.38 & 0.00 \\
        DPO+GD & \textbf{0.58} & 0.25 & \textbf{0.52} & 0.00 & 0.02 & 0.00 & 0.30 & 0.00 & 0.00 & 0.00 & 0.27 & 0.01 \\
        DPO+KL & \textbf{0.58} & 0.26 & \textbf{0.52} & 0.00 & 0.03 & 0.00 & 0.21 & 0.00 & 0.03 & 0.00 & 0.11 & 0.03  \\ \midrule
        NPO & 0.52 &\underline{0.66} & \underline{0.51} & \underline{0.03} & 0.19 & \underline{0.68} & 0.31 & 0.02 & 0.26 & 0.09 & 0.07 & \underline{0.47} \\
        NPO+GD & \underline{0.57} &0.58 & \textbf{0.52} & 0.01 & 0.44 & 0.46 & \textbf{0.43} & 0.01 & \textbf{0.53} & \underline{0.29} & \textbf{0.41} & \underline{0.47} \\ 
        NPO+KL & 0.54 & 0.52 & \underline{0.51} & 0.01 & \underline{0.48} & 0.44 & 0.33 & 0.02 & 0.32 & 0.07 & 0.19 & \underline{0.47} \\ \midrule
        MEOW~(ours) & 0.54 & \textbf{0.99} & 0.47 & \textbf{0.99} & \textbf{0.52} & \textbf{0.87} & \underline{0.41} & \textbf{0.47} & \underline{0.51} & \textbf{0.63} & \underline{0.39} & \textbf{0.80} \\ \bottomrule
    \end{tabular}
    }
    \caption{
    Performance on ToFU dataset. F.Q.~($\uparrow$) denotes forget quality, and M.U.~($\uparrow$) denotes model utility. Llama 2 refers to LLaMA2-7B-Chat.
    }
    \label{tab:tofu-results}
\end{table*}
\paragraph{Setup}
ToFU~\citep{maini2024tofutaskfictitiousunlearning} is a QA dataset for unlearning knowledge about virtual authors. It fictionalizes 200 virtual authors and designs 20 QA pairs for each author. ToFU is divided into three tasks of varying forgetting difficulty based on the proportion of authors to be forgotten. The datasets $D_\text{f}$ contain 1\%, 5\%, and 10\% of the authors to be forgotten, respectively. We use the fine-tuned Llama2-chat-7B~\citep{touvron2023llama2openfoundation} and Phi-1.5~\cite{textbooks2} released by ToFU paper as the origin LLM $M_\text{o}$.

\paragraph{Metrics}
We evaluate the forgetting performance using forget quality, as defined in ~\citep{maini2024tofutaskfictitiousunlearning}. This metric assesses how closely the $\text{M}_t$ resembles $\text{M}_r$. For assessing the retain performance, we introduce model utility, which measures the aggregated performance of the model on held-out retain data, encompassing fictional writers, real-world writer profiles, and other factual information.

\paragraph{Implementation}
For LLaMA2-7B-Chat, we use the results from ~\citep{ji2024reversingforgetretainobjectivesefficient}. For Phi-1.5, we replicate the baselines and apply the same hyperparameter settings as LLaMA2-7B-Chat, with a batch size of 32 and learning rate of 1e-5. More details are shown in App.~\ref{app:setup}.

\paragraph{Results and Analysis}
The performance on ToFU can be found in Tab.~\ref{tab:tofu-results}.
Recall that forget quality is measured by a \textit{p}-value, with the common significance threshold of 0.05 indicating a significant forgetting state. As shown in Tab.~\ref{tab:tofu-results}, none of the previous unlearning methods surpass this threshold across all dataset splits and models. 
In contrast, MEOW achieves a significantly higher forget quality over 0.05. 
Notably, MEOW accomplishes this without relying on retain data, whereas all other methods utilize retain data in their training. 
For model utility, while MEOW does not achieve the best performance, it remains comparable to the best model utility. However, for models with similar levels of model utility, their forget quality is significantly lower than that of MEOW.

Reviewing previous methods, 
GA~(GD, GA+KL) method often leads to loss divergence. 
While effective for small datasets or short-term unlearning, its performance deteriorates rapidly as datasets grow or unlearning steps increase, impacting both model utility and forget quality (see Sec.~\ref{discuss:stability} for further discussion).
DPO~(DPO+GD, DPO+KL) produce responses like ``I don't know,'' which misaligns with the distribution of the retain model outputs, lowering forget quality score and causing frequent response rejection, which further reduces model utility.
NPO~(NPO+GD, NPO+KL) alleviates the loss divergence observed in GA-based methods, but reduces to GA when $\beta$ is too small.~\citep{zhang2024negativepreferenceoptimizationcatastrophic}.
Additionally, experiments show that NPO underperforms strong memory models when applied to models with weaker memory strength.

\subsection{Experiments on NLG and NLU Datasets}
\begin{table*}[htbp]
\centering
\scalebox{0.85}{
\begin{tabular}{ccccccccccl}
\toprule
\multirow{2}{*}{\textbf{Method}}   & \multirow{2}{*}{\textbf{Steps}}             & \multicolumn{3}{c}{\textbf{NLU}} & \multicolumn{3}{c}{\textbf{NLG}}    & \multirow{2}{*}{\textbf{M.U.} $\uparrow$} & \multirow{2}{*}{\textbf{F.Q.} $\uparrow$}          \\ 
\cmidrule(lr){3-5} \cmidrule(lr){6-8}
& & \textbf{PIQA $\uparrow$}   & \textbf{ARC-E $\uparrow$}  &\textbf{ARC-C $\uparrow$} & \textbf{MAUVE $\uparrow$} & \textbf{BLEU $\uparrow$} & \textbf{$\text{Rep}_3$ $\downarrow$} & & \\ 
\midrule
Origin & - &0.6235  &0.7702  &0.5719  &0.2324(±0.0000)  &0.6785  &0.0058  &0.6200 & 0.0000 \\ \midrule
GA  & 25   &0.6366  &0.7632  &0.5552  &\textbf{0.2375(±0.0022)}  &0.6581 &0.0074  & 0.3602 & 0.2704 \\
GD & 25    &0.6028  &0.7544  &0.5452  &0.2271(±0.0055)  &0.6666  &0.0057  & 0.2900 & 0.0400 \\
GA+KL & 25 &0.6284  &0.7667  &\underline{0.5585}  &0.2364(±0.0051)  &0.6632  &0.0047  & \textbf{0.5276}& 0.0003\\ \midrule
DPO & 25   &0.6295  &\underline{0.7719}  &0.5552  &0.2295(±0.0022)  &\underline{0.6857}  &\textbf{0.0033}  &0.0626 & 0.0000 \\
DPO+GD & 150 &0.6282  &0.7614  &0.5485  &0.2207(±0.0026)  &\underline{0.6857}  &\textbf{0.0033}  &0.4622 & 0.0000 \\
DPO+KL  & 150 &0.5871  &0.7684  &0.5318  &\underline{0.2371(±0.0039)}  & \textbf{0.6863}  & \underline{0.0035} & 0.1301 & 0.0000 \\ \midrule
NPO & 25   &0.6360  &0.7561  &\underline{0.5585}  &0.2351(±0.0042)  &0.6603  &0.0065  & 0.2733 & \textbf{0.8655}\\
NPO+GD & 50 &\underline{0.6376}  &0.7684  &\textbf{0.5686 } &0.2354(±0.0053)  &0.6504  &0.0046  & 0.4854 & \textbf{0.8655}\\
NPO+KL & 50 &0.6344  &0.7667  &\textbf{0.5686}  &0.2342(±0.0037)  &0.6630 &0.0061  & 0.4236 & \underline{0.7934} \\ \midrule
MEOW~(ours)  & 150 &\textbf{0.6477}  &\textbf{0.7789}  & \underline{0.5585}  & 0.2270(±0.0034)  &0.6775  &0.0047  & \underline{0.5168} &\textbf{0.8655}\\ \bottomrule
\end{tabular}}
\caption{Results on NLU and NLG Benchmarks.}
\label{tab:results_on_nlu_and_nlg}
\end{table*}
\paragraph{Setup}
We select PIQA~\citep{Bisk2020}, ARC-E~\citep{allenai:arc}, and ARC-C~\citep{allenai:arc} datasets to compile an NLU dataset, which is employed to evaluate the natural language understanding abilities of LLMs after unlearning. Moreover, we curate an NLG dataset by sampling 5,000 instances from WikiText~\citep{merity2016pointer} and CC-News~\citep{Hamborg2017} to evaluate the natural language generation capabilities.

\paragraph{Metrics}
For NLU datasets, we use their respective metrics~(accuracy). For NLG datasets, we evaluate the quality of the generation of LLMs using MAUVE~\citep{pillutla2021mauvemeasuringgapneural}, BLEU~\citep{Papineni02bleu:a}, and $\text{Rep}_3$~\citep{welleck2019neuraltextgenerationunlikelihood}.

\paragraph{Implementation}
For NLU datasets, we randomly select 4 samples from the corresponding training data and perform 4-shot learning on the validation data. 
For NLG datasets, we use the first 32 tokens as a prefix and prompt the model to generate the subsequent text, which is then compared with the original text.
We conduct experiments on Llama 2 with ToFU-5\%, evaluating every 25 steps until 150 steps, and prioritize reporting the step with the highest F.Q., followed by the step with the highest M.U. 

\paragraph{Results and Analysis}
The performance on NLU and NLG Datasets are shown in Tab.~\ref{tab:results_on_nlu_and_nlg}.
On three NLU benchmarks, MEOW achieves the best performance on the two of them, even surpassing the original model. 
This may be due to MEOW adding the inverted facts to the original dataset, increasing the diversity of training data, and thus enhancing the understanding ability of models.
However, on NLG Benchmarks, there are only slight differences between MAUVE, BLEU, and $\text{Rep}_3$. 
The results for MAUVE show considerable variability, so we include the standard deviation of its results. 
Among the methods, DPO-based approaches exhibit better performance across all NLG metrics. Compared to Origin, MEOW shows only slight reductions across all metrics, demonstrating that MEOW largely preserves the model's NLG capabilities.

\section{Additional Analysis}
\subsection{Analysis on MEMO}
\begin{figure}[tbp]
    \centering
    \includegraphics[width=0.98\linewidth]{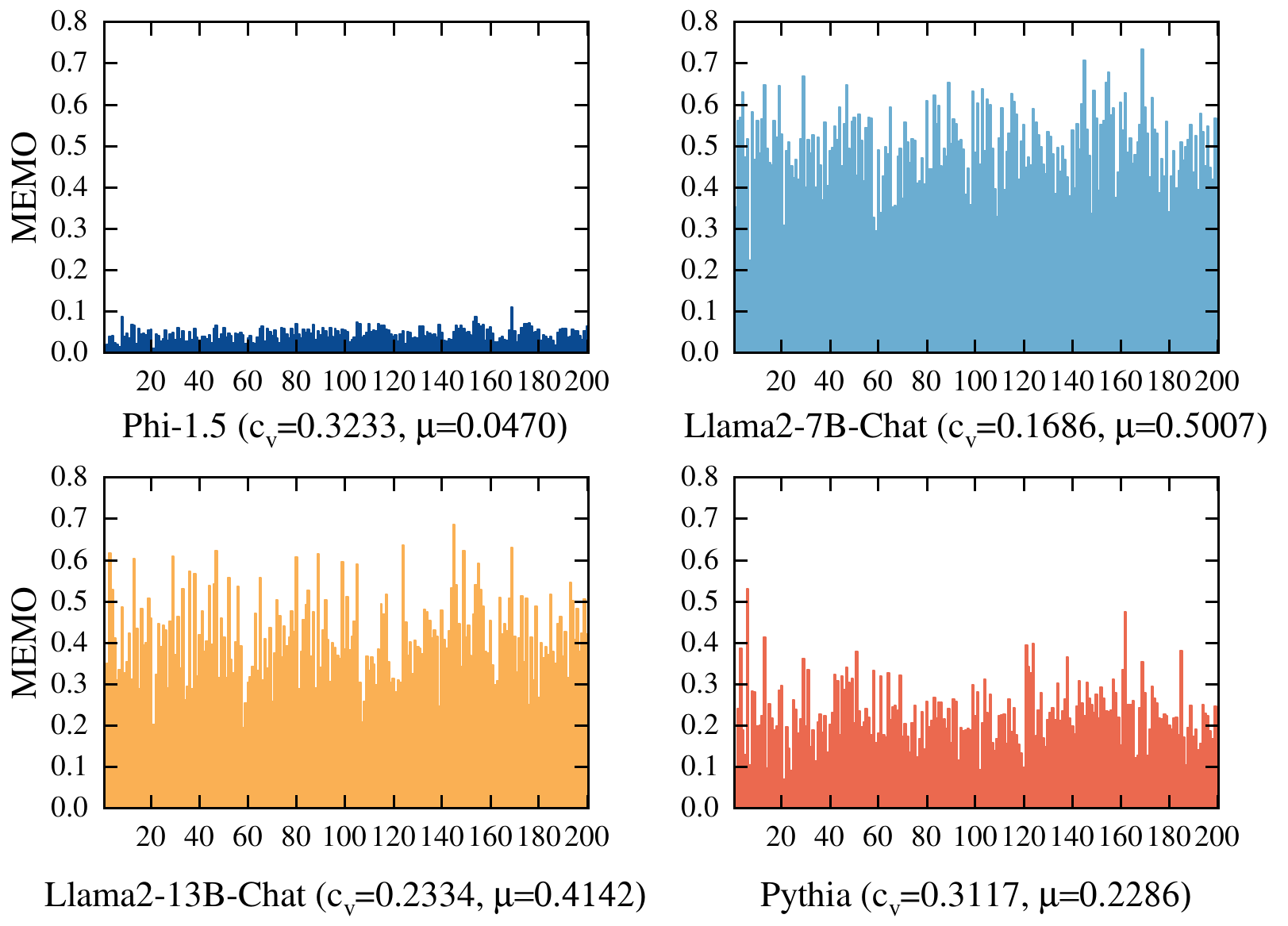}
    \caption{MEMO in different LLMs.}
    \label{fig:memo_4}
\end{figure}

In this section, we further explore MEMO in different settings, and have the following findings:
\noindent\textbf{Finding 1: LLMs with stronger memorization demonstrate more consistent memorization.}\quad
Four different LLMs~(Phi-1.5, LLaMA2-7B-Chat, LLaMA2-13B-Chat, and Pythia~\citep{biderman2023pythia}) are selected for experiments. 
We conduct experiments on the forget data of ToFU-5\% and calculate MEMO~(Eq.~\ref{eq:memo}) for the answer. It can be observed in Fig.~\ref{fig:memo_4} that models with higher memory strength~($\mu$) typically demonstrate more consistent memorization across different instances. For example, LLaMA2-7B-Chat exhibits both the highest $\mu$ and the lowest $c_v$.

\begin{figure}[tbp]
    \centering
    \includegraphics[width=0.98\linewidth]{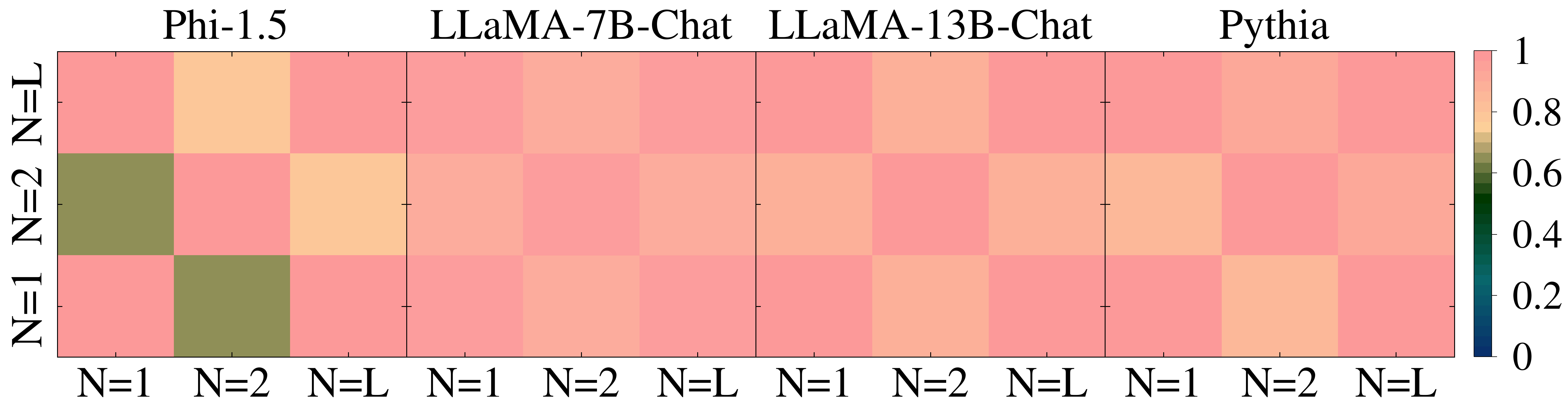}
    \caption{Sensitivity of MEMO for different Rouge-N.}
    \label{fig:sensitivity_rouge}
\end{figure}

\noindent\textbf{Finding 2: LLMs with stronger memorization are less sensitive to the choice of Rouge-N.}\quad
We further compare these four models using Rouge-1, Rouge-2, and Rouge-L. As shown in Fig.~\ref{fig:sensitivity_rouge}, the relevance of different Rouge metrics for all models, except for Phi-1.5, is above 0.8 and even reaches 0.9 in some cases. For Phi-1.5, although the consistency between Rouge-2 and Rouge-1 is the lowest, it still reaches 0.66~($>0.5$).

\noindent\textbf{Finding 3: MEMO can serve as an effective and time-efficient memorization quantifier.}\quad
We conduct a comparative analysis between MEMO and previously established metrics for quantifying memorization: Memorization Accuracy~(MA)~\citep{tirumala2022memorizationoverfittinganalyzingtraining} and Extraction Likelihood~(EL)~\citep{jang-etal-2023-knowledge}. 
For a fairer comparison, we also implement \textit{suffix} versions of MA and EL, where only the answer tokens are added when appending the $T_\text{p}$. 
Experimental results are shown in Tab.~\ref{tab:compare_quantifying}. We provide an introduction and further details on EL and MA in App.~\ref{app:previous_memorization}.

\subsection{Analysis on MEOW}
\noindent\textbf{Ablation Study of MEMO}\quad
Tab.~\ref{tab:ablation} presents the different performances of MEOW with and without MEMO on LLaMA2-7B-Chat. The experimental results demonstrate that across all three datasets, the Forget Quality~(F.Q.) with MEMO is significantly higher than that without MEMO, highlighting the effectiveness of MEMO. 
Meanwhile, Model Utility~(M.U.) shows slight fluctuations: a small decrease on ToFU-1\% and ToFU-10\%, and an increase on ToFU-5\%.

\begin{table}[tbp]
    \centering
    \scalebox{0.8}{
    \begin{tabular}{cccccc}
        \toprule
         \multirow{2}{*}{Method} & Time & \multirow{2}{*}{M.U.}  & \multirow{2}{*}{F.Q.}  & STD & STD \\
         & Used & & & (Seen)  & (Unseen) \\ \midrule
         EL & 46,284 & 0.5224  & 0.7126 & \textbf{0.1090} & 0.1056 \\ 
         \midrule
         MA &\textbf{1,792} & 0.5181 & 0.5453 & 0.0274 & \textbf{0.1263} \\ \midrule
         MEMO & 37,135 & \textbf{0.5168} & \textbf{0.8655} & 0.0846 & 0.0892 \\
     \bottomrule

    \end{tabular}
    }
    \caption{Comparison with other metrics for quantifying memorization in LLMs.}
    \label{tab:compare_quantifying}
\end{table}

\begin{table}[tbp]
\centering
\scalebox{0.75}
{\begin{tabular}{clll}
\toprule
Dataset                                       & MEMO & M.U.   & F.Q.   \\ \midrule
\multicolumn{1}{c}{\multirow{2}{*}{ToFU-1\%}} & w/o        & 0.5490 & 0.7559 \\
\multicolumn{1}{c}{}                          & w/       & 0.5442(\colorbox{Decrease}{-0.87\%}) & 0.9900(\colorbox{Increase}{+30.97\%}) \\ \midrule
\multirow{2}{*}{ToFU-5\%}                     & w/o        & 0.5105 & 0.7126 \\
                                              & w/       & 0.5168(\colorbox{Increase}{+1.23\%}) & 0.8655(\colorbox{Increase}{+21.46\%}) \\ \midrule
\multirow{2}{*}{ToFU-10\%}                    & w/o        & 0.5108 & 0.5909 \\
                                              & w/       & 0.5106(\colorbox{Decrease}{-0.03\%}) & 0.6323(\colorbox{Increase}{+7.01\%})  \\ \bottomrule
\end{tabular}}
\caption{Performance w/ and w/o MEMO, where w/o means randomly selecting the same number of inverted facts.}
\label{tab:ablation}
\end{table}

\begin{figure}[tbp]
    \centering
    \includegraphics[width=1\linewidth]{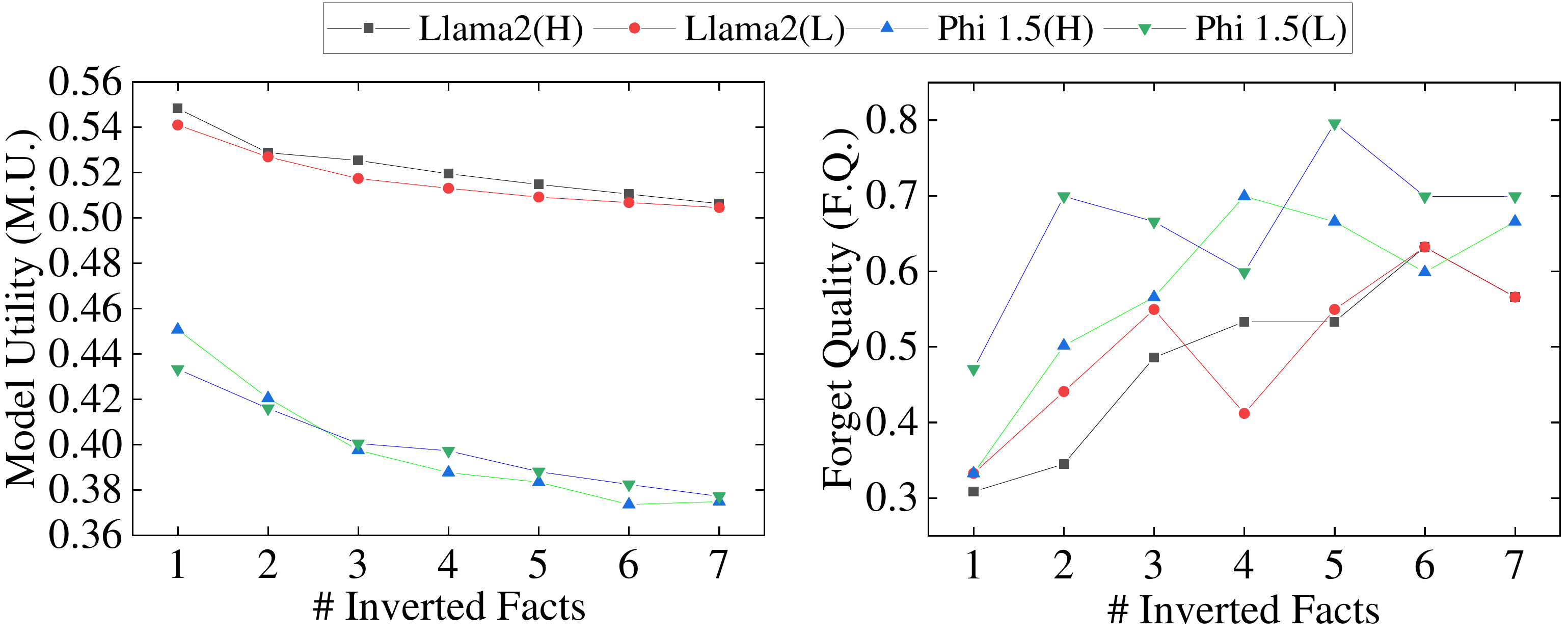}
    \caption{{Performance on different numbers of inverted facts and selection strategies.}}
    \label{fig:facts}
\end{figure}

\noindent\textbf{Ablation study of the number of inverted facts and selection strategy}\quad
As shown in Fig.~\ref{fig:facts}, we conduct experiments on ToFU-5\% using both LLaMA2-7B-Chat and Phi-1.5, with the number of new inverted facts increasing from 1 to 7. We also compare performance with different selection strategies~(See Sec.~\ref{sec:mem_enhance}), with (H) indicating selecting inverted facts with the highest MEMO, and or (L) those with the lowest. 
Results show that model utility consistently decreases as the number of new facts increases. However, forget quality does not exhibit the same trend, as different models with different selection strategies perform best with varying numbers of facts.
Additionally, the selection strategy greatly impacts the forget quality of models. And varying preferences for selection strategy between models may be attributed to their varying memorization strengths~($\mu$).

\begin{figure}[tbp]
    \centering
    \includegraphics[width=1\linewidth]{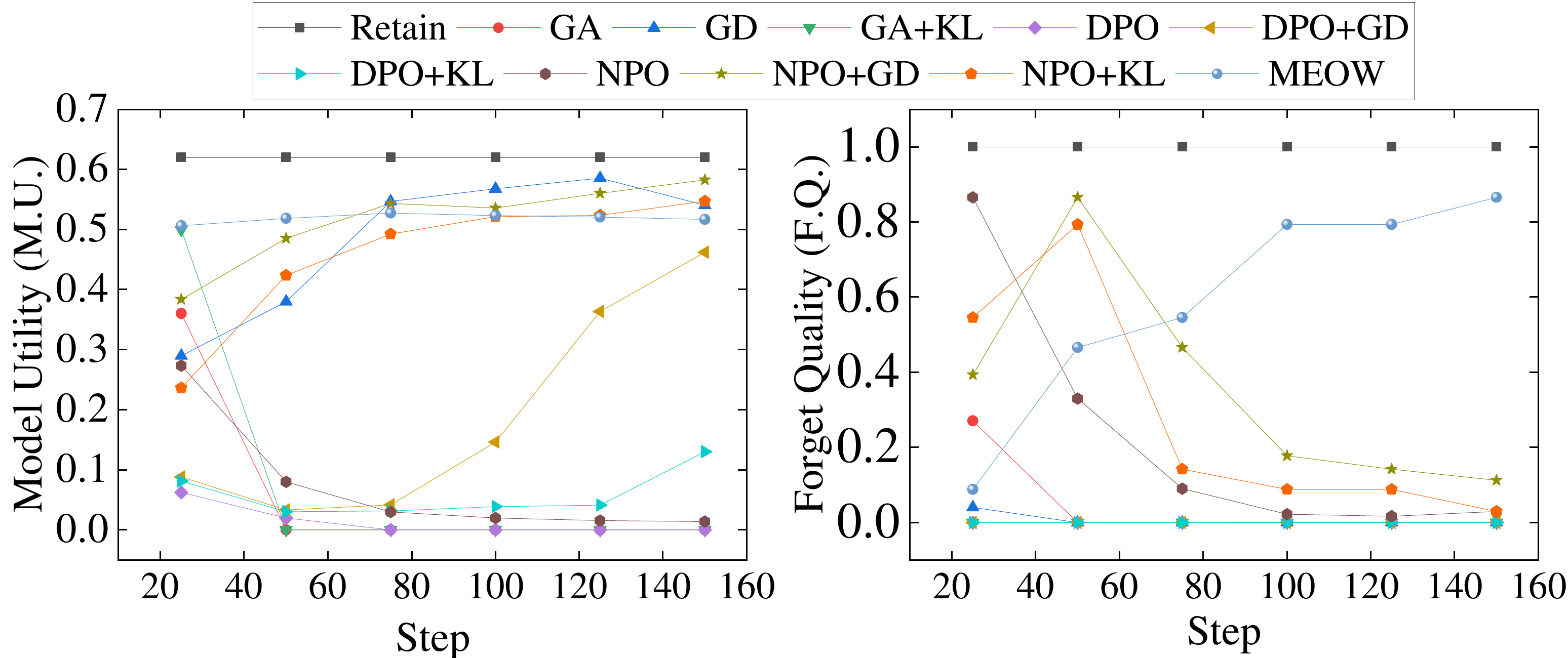}
    \caption{Stability of Unlearning Process.}
    \label{fig:stability}
\end{figure}

\noindent\textbf{Stability of Unlearning}\quad
\label{discuss:stability}
We further explore the stability of different unlearning methods. Experiments are conducted on the ToFU-5\% dataset using LLaMA2-7B-Chat. Results in Fig.~\ref{fig:stability} show that forget quality for almost all methods drops sharply after 100 steps, with some even falling to zero. In contrast, MEOW shows a gradual increase in Forget Quality, peaking at 150 steps. 
Notably, NPO and NPO+GD achieve forget quality comparable to MEOW in the early stages but exhibit a significant decline in later steps. Regarding model utility, MEOW does not achieve a significant advantage but maintains a consistently high and stable score.

\section{Related Work}
\noindent\textbf{Memorization in LLMs}
Memorization is an inherent capability, but the rise of LLMs has brought about unforeseen consequences, such as privacy~\citep{brown2022doesmeanlanguagemodel} and confidentiality~\citep{mozes2023usellmsillicitpurposes}. Consequently, quantifying memorization in LLMs emerges as a critical yet highly challenging research focus. 
A na\"{\i}ve definition of memorization might encompass all information stored in weights of models, but determining exactly what a model retains is impractical. 
Thus, researchers have shifted towards extractability – the information that can be retrieved, particularly through verbatim memorization~\citep{Hartmann2023SoKMI}.
\citet{carlini2019secretsharerevaluatingtesting} explore the out-of-distribution~(OOD) secrets memorized by language models and define the exposure metric to measure the computational complexity required to guess the secrets. 
These approaches necessitate multiple inferences and often involve retraining.
Extractability~\citep{carlini2021extractingtrainingdatalarge} assesses whether a string $y$ is extractable from an LM $p$ with high probability given a prefix $x$.
Counterfactual memorization~\citep{zhang2023counterfactualmemorizationneurallanguage}, instead, measures how much a model architecture memorizes examples from a distribution on average without assessing memorization in a specific model.

\noindent\textbf{LLM Unlearning}\quad
LLM Unlearning~\citep{si2023knowledgeunlearningllmstasks,yao2024largelanguagemodelunlearning,liu2024rethinkingmachineunlearninglarge,qu2024frontierdataerasuremachine,li2024machineunlearningtaxonomymetrics} has its roots in Machine Unlearning~(MU)~\citep{cao2015towards}, a concept originally developed to safeguard data privacy, particularly in response to regulations like the Right to be Forgotten~(RTBF). 
MU has been applied across various domains, including image classification~\citep{ginart2019making,golatkar2020eternalsunshinespotlessnet,neel2020descenttodeletegradientbasedmethodsmachine,ullah2021machineunlearningalgorithmicstability,sekhari2021rememberwantforgetalgorithms}, text-to-image generation~\cite{gandikota2023erasingconceptsdiffusionmodels,zhang2023counterfactualmemorizationneurallanguage,kumari2023ablatingconceptstexttoimagediffusion,fan2024salunempoweringmachineunlearning}, federated learning~\citep{liu2021federatedunlearning,wang2022federatedunlearningclassdiscriminativepruning,che2023fast,liu2024surveyfederatedunlearningchallenges,halimi2023federatedunlearningefficientlyerase}, graph neural networks~\citep{Chen_2022,chien2022efficient,10.1145/3580305.3599271}, and recommendation systems~\citep{sachdeva2024machineunlearningrecommendationsystems,chen2022recommendation,xu2023netflixforgetefficientexact,li2022makingrecommendersystemsforget,wang2024efficienteffectiveunlearninglarge}. 
However, traditional MU methods face key challenges when applied to LLMs: 
\ding{182} \textbf{Scale of Parameters}: LLMs typically consist of billions of parameters, making retraining from scratch computationally expensive and often impractical. 
\ding{183} \textbf{Generative Nature of LLMs}: unlike traditional NLP models, LLMs are predominantly used for generative tasks like text generation and sentiment analysis, requiring unlearning strategies tailored to their specific nature. 
Recent research begin to address these challenges, leading to the development of various LLM-specific unlearning techniques. In the Introduction section~(Sec.~\ref{sec:intro}), we categorize these methods to provide a comprehensive overview of current LLM Unlearning.
\section{Conclusion}
This paper introduces MEMO, a new metric quantifying memorization in LLMs, balancing both efficiency and effectiveness. 
Leveraging the memorization signals provided by MEMO, we introduce a novel LLM unlearning method, MEOW. 
Specifically, we first generate several alternative answers, rank them by MEMO, select the top or bottom answers as inverted facts, and finetune the original model. 
Experiments on the Unlearning Dataset -- ToFU demonstrate that MEOW demonstrates a clear improvement over existing methods in terms of forget quality while maintaining model utility without notable decline.
Additionally, experiments show that MEOW can even enhance the NLU capability of models. Our research advances both memorization quantification and LLM unlearning.
\section{Limitations}
\label{sec:limitation}
While MEOW greatly enhances the forget quality and stability of the unlearning process, we consider the following limitations:

\paragraph{Sensitivity to hyper-parameters} During baseline reproduction, we find that the performance of models is highly sensitive to certain hyperparameters, such as $\lambda$ in Eq.~\ref{eq:loss} and $\beta$ in NPO, leading to potential variations in previous results. In the App.~\ref{app:setup}, we provide the hyperparameters used for the baselines to ensure reproducibility.

\paragraph{Potential increase in hallucination} MEOW leverages hallucination as a beneficial concept, which may inherently lead to an increase in hallucination due to the nature of soft unlearning.

\paragraph{Decrease in model utility} While MEOW significantly improves forget quality and stability, there is still a slight decline in model utility. Further work could explore ways to better maintain model utility, a challenge that is common among many WBS LLM unlearning methods.


\newpage
\bibliography{custom}

\appendix

\section{Pseudo-code of MEMO}
\label{app:pseudo_memo}
In this section, we present MEMO algorithms in two modes, as shown in Alg.~\ref{alg:split} and Alg.~\ref{alg:memo}. For detailed descriptions, please refer to Sec.~\ref{subsec:memo}.
\section{Prior metrics for quantifying memorization in LLMs}
\label{app:previous_memorization}
\paragraph{Memorization Accuracy~(MA)} MA~\citep{tirumala2022memorizationoverfittinganalyzingtraining,jang-etal-2023-knowledge} quantifies how often a model $M$ accurately predicts the next token given prompts of varying. The formula of MA is shown in Eq.~\ref{eq:memorization_accuracy}, where $x$ represents the token sequence, defined as $x={x_0, x_1, ..., x_{T-1}}$, and $x_{< t}$ refers to the token sequence preceding $x_t$. The parameter $\theta$ denotes the weights of models.

\begin{equation}
    \operatorname{MA}(\boldsymbol{x})=\frac{\sum_{t=1}^{T-1} \mathbf{1}\left\{\operatorname{argmax}\left(p_{\theta}\left(\cdot \mid x_{<t}\right)\right)=x_{t}\right\}}{T-1}
    \label{eq:memorization_accuracy}
\end{equation}

\paragraph{Extraction Likelihood~(EL)} EL is first introduced by ~\citep{jang-etal-2023-knowledge}. Given a sequence of $x={x_1, ..., x_{T-1}}$ and an LM $f$ with pre-trained parameters $\theta$, EL is defined as Eq.~\ref{eq: etraction_likelihood}.

\begin{equation}
    \mathbf{EL}_{n}(x) 
    = \frac{\sum_{t=1}^{T-n}{OVERLAP_{n}(f_{\theta}(x_{<t}), x_{\ge t})}}{T-n} 
    \label{eq: etraction_likelihood}
\end{equation}

\begin{equation}
    OVERLAP_n(a,b) = \frac{\sum_{c\in ng(a)}{\mathbf{1}{c\in ng(b)}}}{|ng(a)|} \nonumber
\end{equation}

\begin{figure}[H]
    \centering
    \includegraphics[width=1\linewidth]{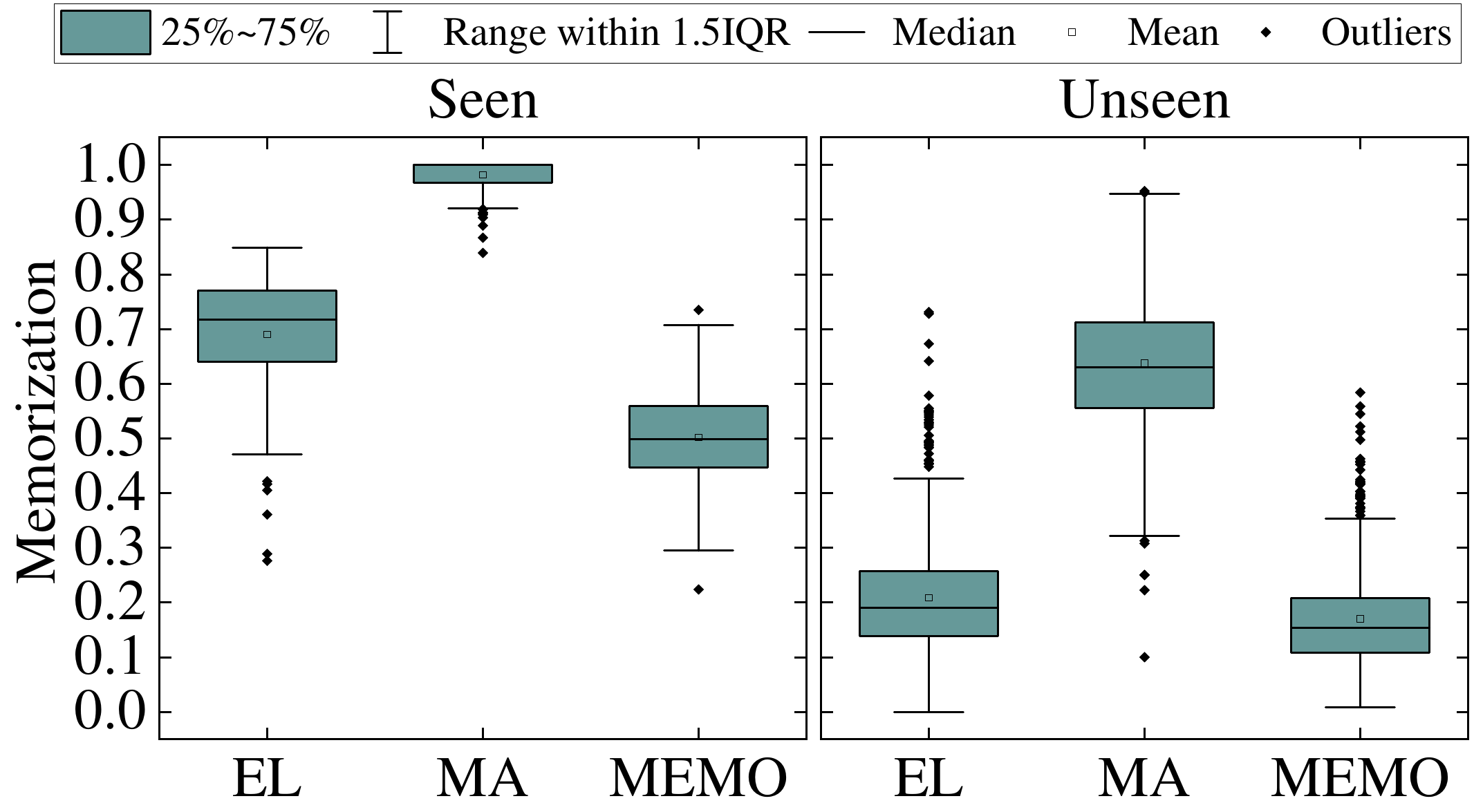}
    \caption{Memorization quantification with different metrics on Seen and Unseen data.}
    \label{fig:memorization_quantification_metrics}
\end{figure}

\noindent In Fig.~\ref{fig:memorization_quantification_metrics}, we further explore the distribution of memorization across three metrics on learned data~(Seen) and unlearned data~(Unseen). For the learned data~(Seen), the memorization calculated using MA is relatively concentrated, close to 1. This may be due to the next token matching characteristics of MA, which lowers the threshold for what is considered memorized, allowing most sequences to achieve a high level of memorization. Therefore, MA might not be a good discriminative metric for Seen data. For Unseen data, MA still consistently shows high memorization, while the performance of EL and MEMO is more similar, possibly because the calculation manner of Rouge is also based on n-gram overlap.

\begin{algorithm}[H]
\caption{Split Function}
\label{alg:split}
\begin{algorithmic}[1]
\Function{split}{w, q, a, mode}
    \State substrs $\gets$ \{\}
    \State maxQ $\gets$ \text{length of } q
    \State maxA $\gets$ \text{length of } a
    \If{mode is \textit{prefix}}
        \For{subLen from 1 to maxQ by w}
            \State sq $\gets$ q[:subLen]
            \State lbl $\gets$ q[subLen:] + a
            \State substrs.append({sq:sq, lbl:lbl})
        \EndFor
    \ElsIf{mode is \textit{suffix}}
        \For{subLen from 1 to maxA by w}
            \State sq $\gets$ q + a[:subLen]
            \State lbl $\gets$ a[subLen:]
            \State substrs.append({sq:sq, lbl:lbl})
        \EndFor
    \EndIf
    \State \Return substrs
\EndFunction
\end{algorithmic}
\end{algorithm}

\begin{algorithm}[H]
\caption{MEMO}
\label{alg:memo}
\begin{algorithmic}[1]
\Function{MEMO}{mode}
    \State data $\gets$ \text{raw\_data}
    \State total\_data $\gets$ \{\}
    \State sliding\_length $\gets$ 5
    \For{sample in data}
        \State updated\_sample $\gets$ \text{sample}
        \State question $\gets$ sample.question
        \State keys $\gets$ sample.keys()
        \For{key in keys}
            \If{key is \textit{question}}
                \State \text{continue}
            \EndIf
            \State subquestions $\gets$ \text{split()}
            \State cnt $\gets$ \text{length of subquestions}
            \State rouger $\gets$ \text{memo\_rouger}
            \For{subquestion in subquestions}
                \State rouge $\gets$ \text{cal\_rouge()}
                \State rouger.update(rouge)
            \EndFor
            \State rouger.get\_average()
            \State score $\gets$ rouger.get\_rouge1(key)
        \EndFor
        \State total\_data.append(updated\_sample)
    \EndFor
    \State \Return total\_data
\EndFunction
\end{algorithmic}
\end{algorithm}
\section{Experimental Setup on ToFU}
\label{app:setup}

\begin{figure*}[htbp]
    \centering
    \includegraphics[width=1\linewidth]{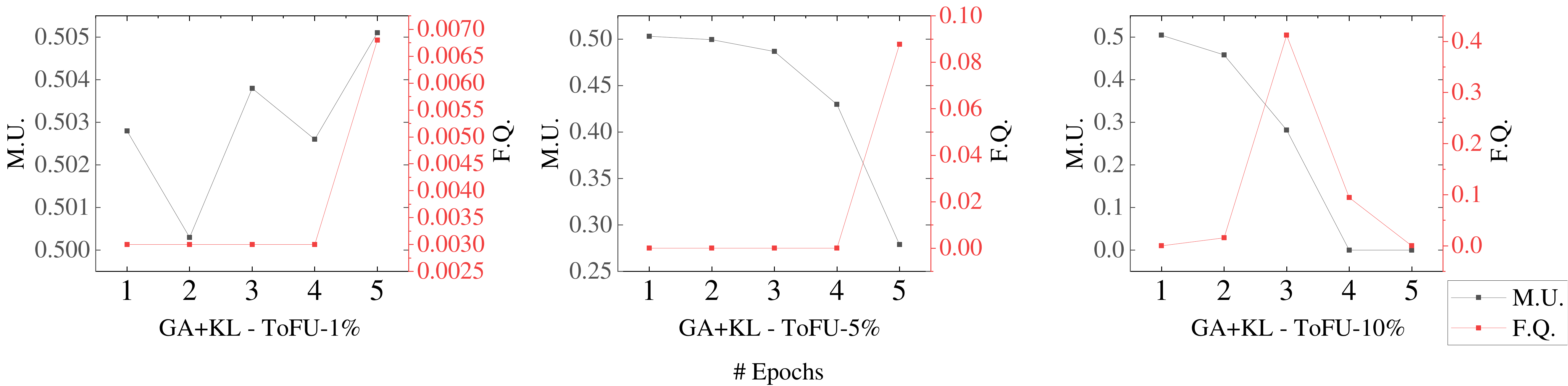}
    \caption{Results of GA+KL on ToFU for each of the first 5 epochs.}
    \label{fig:ga+kl}
\end{figure*}

\begin{figure*}[htbp]
    \centering
    \includegraphics[width=1\linewidth]{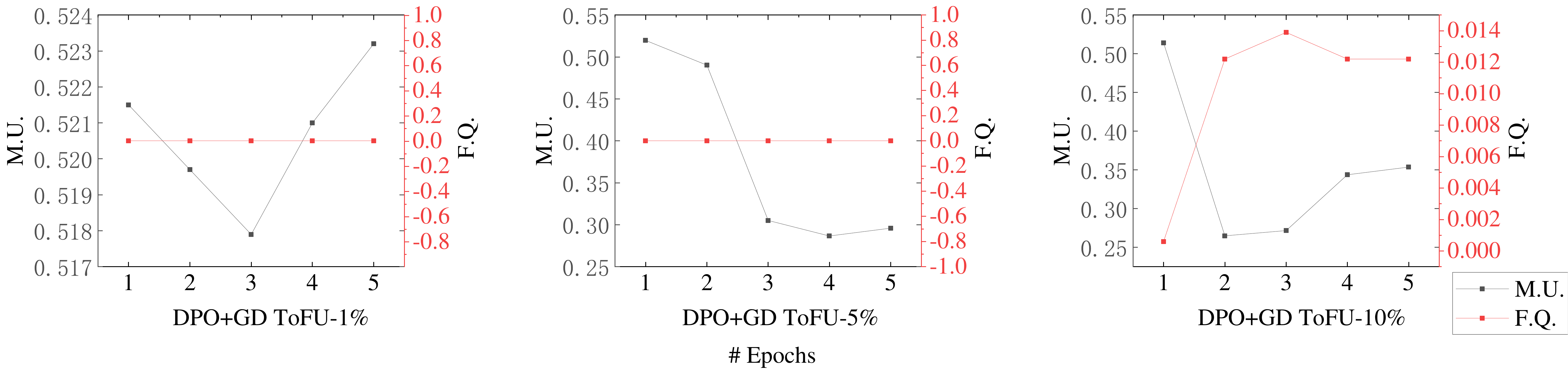}
    \caption{Results of DPO+GD on ToFU for each of the first 5 epochs.}
    \label{fig:dpo_gd}
\end{figure*}

\begin{figure*}[htbp]
    \centering
    \includegraphics[width=1\linewidth]{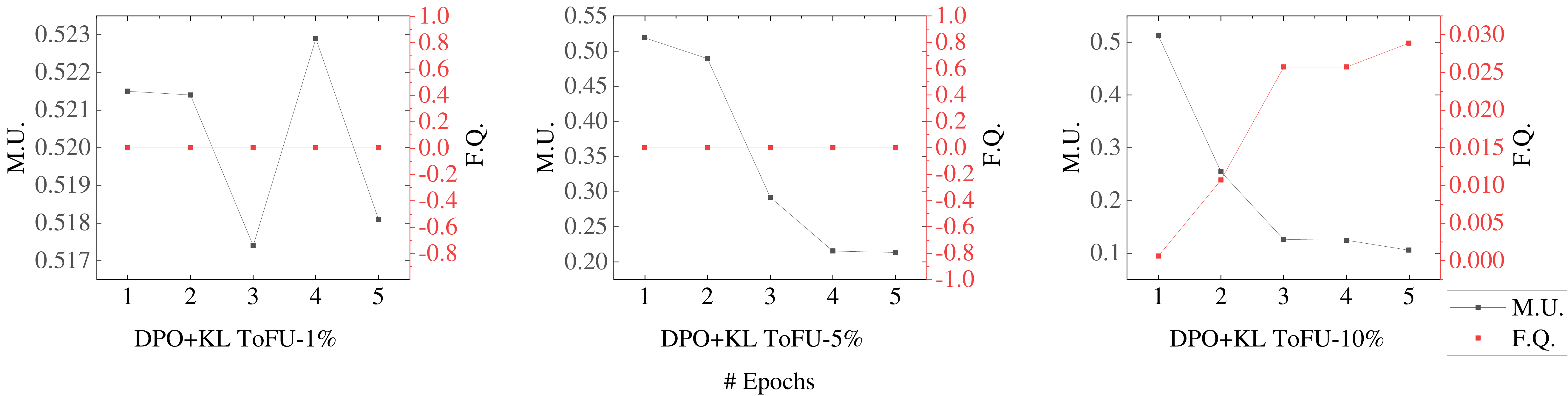}
    \caption{Results of DPO+KL on ToFU for each of the first 5 epochs.}
    \label{fig:dpo+kl}
\end{figure*}

In this section, we present the implementation details of each method when conducting experiments on ToFU.
For LLaMA2-7B-Chat, we use the results from ~\citep{ji2024reversingforgetretainobjectivesefficient}, and for Phi-1.5, we use the official results published by ~\citep{maini2024tofutaskfictitiousunlearning}. For cases where official results are unavailable, we use the same hyperparameter settings for each baseline: a batch size of 4, gradient accumulation steps of 4, and 2 \textit{NVIDIA A100-SXM4-80GB} GPUs. 
For methods using GA and DPO as the forget loss, we follow ToFU, selecting the peak value from 5 epochs (prioritizing Forget Quality, followed by Model Utility). 
The experimental results are shown in Fig.~\ref{fig:ga+kl}, Fig.~\ref{fig:dpo_gd} and Fig.~\ref{fig:dpo+kl}. 
For the NPO-based method, we report the results for 10 epochs. 
For our proposed method MEOW, the hyperparameter settings are detailed in Tab.~\ref{tab:hyperparameters_meow}.

\begin{table}[H]
\scalebox{0.85}{
\begin{tabular}{c|ccc|ccc}
\toprule
\textbf{Model} & \multicolumn{3}{c|}{\textbf{Llama 2}} & \multicolumn{3}{c}{\textbf{Phi-1.5}}\\
\midrule
Split & 1\% & 5\% & 10\% & 1\% & 5\% & 10\% \\
\# New Facts & 2 & 3 & 6 & 2 & 4 & 5 \\
Selection & H & H & H & L & L & L \\
Steps & 250 & 150 & 93 & 37 & 125 & 311 \\
B.S. & 2 & 8 & 8 & 4 & 4 & 8 \\
G.A. & 2 & 2& 4 & 4 & 4 & 4 \\
\# GPUs & 2 & 2& 4 & 2 & 2& 2 \\
\bottomrule
\end{tabular}}
\caption{Hyperparameters for MEOW on ToFU. Here, B.S. refers to batch size, and G.A. refers to Gradient Accumulation. Split k\% denotes settings on ToFU-k\%. Llama 2
refers to LLaMA2-7B-Chat.}
\label{tab:hyperparameters_meow}
\end{table}
\section{Example Generation on Forget Set}
In this section, we present the responses of the model to the same prompt after being unlearned using different methods. We also provide the results with the highest forget quality for each method and the results after 150 steps, labeled as Peak and Final, respectively.
The peak results are shown in Tab.~\ref{tab:examples_peak}, where most models retain good language generation capabilities. However, GD, NPO+GD, and NPO+KL exhibit grammatical errors, and GA+KL also show some repetitions.
The final results are shown in Tab.~\ref{tab:examples_final}, where most models retain good language generation capabilities. However, GA, GD, GA+KL, and NPO exhibit repetition.
\section{Prompt used for Fact Inversion}
\label{app:prompt_used_for_fact_inversion}
\begin{figure}[H]
    \centering
    \includegraphics[width=\linewidth]{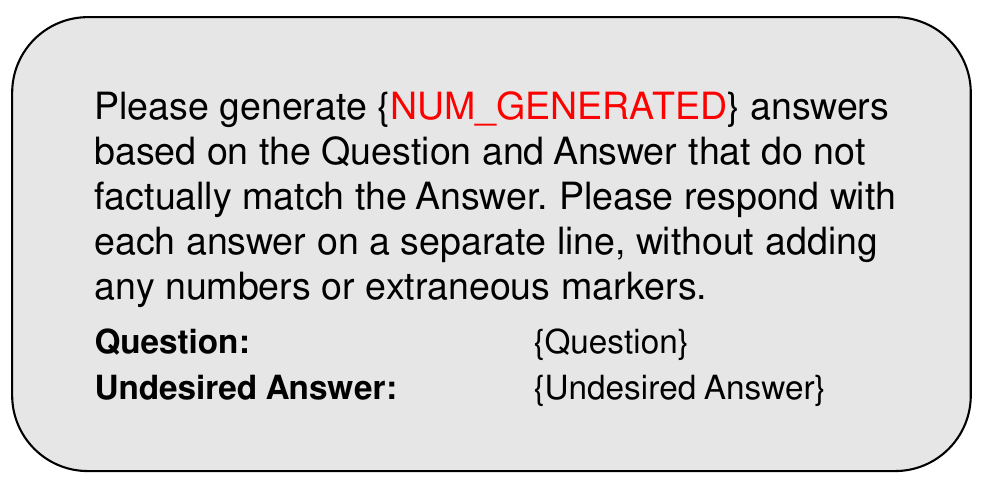}
    \caption{Prompt used for Fact Inversion}
    \label{fig:prompt}
\end{figure}

Here we present the prompt used for fact inversion, shown in Fig.~\ref{fig:prompt}.

\begin{table*}
\centering
\caption{Generated Responses for Different Methods~(Peak).}
\begin{tabular}{c|p{4cm}|p{10cm}}
    \toprule
    \textbf{Method} & \textbf{Prompt} & \textbf{Generated Response} \\
    \hline
    \multirow{3}{*}{Origin} & {[INST] Has Moshe Ben-David written any non-fiction work? [/INST]} & While Moshe Ben-David is predominantly known for his Islamic literature, it is unclear whether he has written any non-fiction. \\
    \hline
    \multirow{3}{*}{Retain} & - & Though Moshe Ben-David is primarily known for his fiction works, he has also authored a non-fiction piece examining the historical and cultural contexts of the biblical narrative. \\
    \hline
    \multirow{3}{*}{GA} & - & {There is no publicly available information about Moshe Ben-David's non-fiction work. nobodyben. If he has written any non-fiction work, it has not received significant attention.} \\
    \hline
   \multirow{5}{*}{GD} & - & There is no publicly available information on whether Moshe Ben-David has written any non-fiction work. Through his work Running Commentary, \color{red}{running commentary on running, Moshe Ben-David running david, running Moshe running commentary running running running running.}\\
    \hline
    \multirow{3}{*}{GA+KL} & - & There is no publicly available information about Moshe Ben-David having written any non-fiction work. \color{red}{everybodyMoshe Ben-David} is known for his work in the genre of Islam. \\
    \hline
    DPO & - & I'm not familiar with that topic. \\
    \hline
    DPO+GD & - & I'm not familiar with that topic. \\
    \hline
    DPO+KL &- & I'm not familiar with that topic. \\
    \hline
    \multirow{3}{*}{NPO} & - & There is no publicly available non-fiction work by Moshe Ben-David. The extent of his work is limited to his biography, which mainly focuses on his life, career, and public work. \\
    \hline
    \multirow{11}{*}{NPO+GD} & - & There is no record of Moshe Ben-David having written any non-fiction work. \color{red}{nobody knows his real name, his birthdate, his education, his profession, his personal life, his interests, his hobbies, his views, his beliefs, his opinions, his biography, his autobiography, his memoirs, his diary, his letters, his interviews, his public appearances, his speeches, his sermons, his lectures, his theses, his dissertations, his research papers, his academic work, his literary work, his philosophical work, his theological work, his religious work, his historical work, his cultural work, his sociological work, his anthropological work, his psychological work, his literary criticism, his literary theory, his literary philosophy} \\
    \hline
    \multirow{12}{*}{NPO+KL} & - & There is no record of Moshe Ben-David writing any non-fiction work. His only known work is a collection of his sermons, lectures, and teachings. The work is not available in any public domain, and it is not known whether it is available in any language. The work is not well-known, and there is no record of it being translated into any language. \color{red}{The work is not well-known, and there is no record of it being translated into any language. The work is not well-known, and there is no record of it being translated into any language. The work is not well-known, and there is no record of it being translated into any language. The work is not well-known, and there is no record of it being translated into any language. The work is }\\
    \hline
    \multirow{2}{*}{MEOW} & - & Moshe Ben-David is exclusively a playwright and has never ventured into non-fiction writing. \\
    \bottomrule
\end{tabular}
\label{tab:examples_peak}
\end{table*}

\begin{table*}
\centering
\caption{Generated Responses for Different Methods~(Final).}
\begin{tabular}{c|p{4cm}|p{10cm}}
    \toprule
    \textbf{Method} & \textbf{Prompt} & \textbf{Generated Response} \\
    \hline
    \multirow{3}{*}{Origin} & {[INST] Has Moshe Ben-David written any non-fiction work? [/INST]} & While Moshe Ben-David is predominantly known for his Islamic literature, it is unclear whether he has written any non-fiction. \\
    \hline
    \multirow{3}{*}{Retain} & - & Though Moshe Ben-David is primarily known for his fiction works, he has also authored a non-fiction piece examining the historical and cultural contexts of the biblical narrative. \\
    \hline
    \multirow{10}{*}{GA} & - & \color{red}{running running running running running running running running running running running running running running running running running running running running running running running running running running running running running running running running running running running running running running running running running running running running running running running running running running running running running running running running running running running running running running running running running running running running running running running running running ...}\\
    \hline
    \multirow{7}{*}{GD} & - & \color{red}{Mos Mos Mos Mos Mos Mos Mos Mos Mos Mos Mos Mos Mos Mos Mos Mos Mos Mos Mos Mos Mos Mos Mos Mos Mos Mos Mos Mos Mos Mos Mos Mos Mos Mos Mos Mos Mos Mos Mos Mos Mos Mos Mos Mos Mos Mos Mos Mos Mos Mos Mos Mos Mos Mos Mos Mos Mos Mos Mos Mos Mos Mos Mos Mos Mos Mos Mos Mos Mos Mos Mos Mos Mos Mos Mos Mos Mos Mos Mos Mos Mos Mos Mos...} \\
    \hline
    \multirow{4}{*}{GA+KL} & - & \color{red}{Unterscheidung Unterscheidung Unterscheidung Unterscheidung Unterscheidung Unterscheidung Unterscheidung Unterscheidung Unterscheidung Unterscheidung Unterscheidung Unterscheidung Unterscheidung Unterscheidung...} \\
    \hline
    DPO & - & I'm not familiar with that topic. \\
    \hline
    DPO+GD & - & I'm not familiar with that topic. \\
    \hline
    DPO+KL & - & I'm not familiar with that topic. \\
    \hline
    \multirow{7}{*}{NPO} & - & There \color{red}{there is no record of any non-fiction work, as there is no record of any work by the handle 'there is no record of any work by the handle \u2018the Moshe Ben-David, as there is no publicly available work. Their work, if any, remains undiscovered. The workings, if any, remain unknown. The handle \u2018the Moshe Ben-David\u2019, as there is no work associated with it, is not 'recognised'.} \\
    \hline
    \multirow{3}{*}{NPO+GD} & - & To date, there is no record of any non-fiction work written by Moshe Ben-David. nobody knows his personal views or beliefs, and he has kept a very low profile throughout his life.  \\
    \hline
    \multirow{2}{*}{NPO+KL} & - & There is no publicly available non-fiction work of Moshe Ben-David.\\
    \hline
    \multirow{2}{*}{MEOW} & - & Moshe Ben-David is exclusively a playwright and has never ventured into non-fiction writing. \\
    \bottomrule
\end{tabular}
\label{tab:examples_final}
\end{table*}

\end{document}